\documentclass[journal]{IEEEtran}
\IEEEoverridecommandlockouts

\usepackage{cite}
\usepackage{amsmath,amssymb,amsfonts}
\usepackage{algorithmic}
\usepackage{graphicx}
\usepackage{booktabs}
\usepackage{textcomp}
\usepackage{xcolor}

\def\BibTeX{{\rm B\kern-.05em{\sc i\kern-.025em b}\kern-.08em
    T\kern-.1667em\lower.7ex\hbox{E}\kern-.125emX}}

\begin{document}

\title{PCGD: Physics-Guided Conditional Graph Diffusion for TCAD Device Simulation}

\author{Yihan~Zhang, 
        Zhiteng~Zhang, 
        Kun~Chen, 
        and~Chen~Wang
\thanks{Y. Zhang, Z. Zhang, K. Chen, and C. Wang are with the College of Integrated Circuits \& Micro-Nano Electronics, Fudan University, Shanghai 200433, China.}%
\thanks{K. Chen and C. Wang are also with the Shanghai Integrated Circuit Manufacturing Innovation Center, Shanghai 201203, China (Corresponding authors: K. Chen; C. Wang, e-mail: chenk18@fudan.edu.cn; chen\_w@fudan.edu.cn).}%
}

\maketitle

\begin{abstract}
Technology computer-aided design (TCAD) semiconductor device simulation is fundamentally constrained by the high computational cost of iteratively solving coupled drift-diffusion equations. Existing ML surrogates either reduce internal physics to macroscopic scalar regressions, or rely on single-step mappings that lack the iterative refinement required to resolve stiff, coupled fields. To address this, we introduce PCGD, a Physics-Guided Conditional Graph Diffusion framework operating natively on unstructured TCAD meshes to predict coupled electrostatic and carrier density fields. PCGD employs a Condition-Aware MeshGraphNet denoiser that explicitly injects boundary conditions and device structure context via global cross-attention. By augmenting data-driven denoising with a physics-guided hybrid objective that integrates exponent-free quasi-Fermi gradient matching with noise-aware PDE residuals, PCGD progressively enforce physical constraints in the iterative diffusion trajectory. This strategy successfully bypasses the numerical instabilities typical of stiff drift-diffusion equations. Evaluated on a challenging mixed PN/MOS benchmark, PCGD significantly outperforms deterministic one-step regression (1.207\% error) and local diffusion (1.585\% error) baselines by achieving a sub-percent mean relative field error of 0.835\%, while concurrently reducing maximum PDE residual errors by nearly three orders of magnitude compared to pure diffusion. It also transfers robustly to unseen SOI topologies (0.815\% error) via LoRA adaptation, using 5.30$\times$ less data and 14.34$\times$ fewer parameters than full fine-tuning. Ultimately, PCGD bridges the computational efficiency of generative surrogates with the rigorous physical fidelity of traditional TCAD, unlocking highly scalable, field-level analysis for robust device engineering.
\end{abstract}

\begin{IEEEkeywords}
Diffusion models, Graph neural networks, Physics-informed learning, Semiconductor device simulation, TCAD
\end{IEEEkeywords}

\section{Introduction}

Technology computer-aided design (TCAD) is fundamental to predictive semiconductor device analysis. It solves coupled electrostatic and carrier-transport equations across complex device geometries, doping profiles, and boundary conditions \cite{Selberherr1984TCAD,scharfetterLargesignalAnalysisSilicon1969}. However, this high fidelity comes at a substantial computational cost, creating a bottleneck for design-space exploration and inverse optimization. While machine-learning surrogates aim to accelerate this process, most conventional models operate as "black boxes". These models directly map device parameters to macroscopic terminal responses, such as $I\text{--}V$ curves \cite{wangArtificialNeuralNetworkBased2021,tungNeuralNetworkBasedBSIM2023,mamunComprehensiveReviewMachine2025}. Although sufficient for circuit-level modeling, these lumped metrics obscure the internal physics that is essential for device optimization. To diagnose microscopic failures and guide spatial topology optimization, it is imperative to break this black box and directly predict high-resolution field-level states—such as peak electric fields, leakage paths, and depletion boundaries \cite{Selberherr1984TCAD,stettlerIndustrialTCADModeling2021}.

Recent graph-based approaches have been introduced on unstructured device meshes to predict these internal fields \cite{jangTCADDeviceSimulation2023,fanGraphAttentionNetworkBased2025}. However, generating a complete, fully coupled device state remains fundamentally challenging owing to the extreme stiffness and strong exponential nonlinearity of semiconductor physics. The electrostatic potential and carrier densities are tightly coupled through drift-diffusion equations, with carrier concentrations spanning dozens of orders of magnitude \cite{caoPhysicsInformedNeuralNetworks2023,rigantiDDNetUnifiedPhysicsInformed2025}. Additionally, steep spatial gradients emerge abruptly at PN junctions and material interfaces. Existing single-step deterministic regression models struggle with these harsh physical constraints and abrupt spatial variations \cite{jangTCADDeviceSimulation2023,fanGraphAttentionNetworkBased2025}. When these models attempt to simultaneously balance global field accuracy against sensitive local conservation laws, they encounter gradient conflicts. This typically hinders the training process, leading to over-smooth predictions that fail to resolve critical high-frequency spatial details \cite{baldanPHYSICSVSDISTRIBUTIONS2026,lippePDERefinerAchievingAccurate2023}.

Overcoming this multiphysics challenge requires a mechanism that separates macroscopic field construction from microscopic residual enforcement. Generative diffusion models offer a well suited solution. The iterative denoising trajectory elegantly decomposes the generation task: the model first establishes the global topological distribution during early high-noise stages, and subsequently resolves sharp local gradients during later low-noise stages \cite{lippePDERefinerAchievingAccurate2023}. More importantly, this timeline provides a natural schedule for progressively enforcing physics guidance. Highly nonlinear PDE residuals can be suppressed when the predicted state is unreliable, and then progressively activated as the prediction approaches the valid physical manifold. This effectively bypasses the gradient divergence that destabilizes standard training. Despite this immense potential, the application of diffusion to solve stiff, nonlinearly coupled drift-diffusion systems directly on irregular TCAD graphs remains unexplored.

To bridge this gap, we introduce PCGD, a Physics-guided Conditional Graph Diffusion framework for semiconductor TCAD surrogates. PCGD formulates coupled-field prediction as conditional generative denoising directly on native TCAD meshes. It jointly generates $(\psi,\log n,\log p)$ on the unstructured device graph, injects bias conditions and device context globally through cross-attention tokens, and progressively introduces physics guidance via a noise-aware, adaptively scaled PDE residual loss and an exponent-free edge-force regularizer.

The core contributions of this work are threefold:
\begin{itemize}
    \item \textbf{Mesh-Native Graph Diffusion Framework:} We propose a graph-diffusion framework that operates directly on irregular TCAD finite-volume meshes. This representation avoids rasterization and preserves the native connectivity used for field generation and residual evaluation.
    \item \textbf{Global Conditioning via Cross-attention Message Passing:} PCGD introduces a Condition-Aware MeshGraphNet that encodes contact biases and device context as conditioning tokens. By injecting these tokens globally through cross-attention, this architecture allows every mesh node to access terminal and structural information directly, without relying solely on long-range local message passing.
    \item \textbf{Hybrid Physics-Guided Training with Noise-Aware Adaptation:} PCGD leverages the iterative diffusion trajectory to enforce physical constraints progressively. It utilizes a stable, exponent-free edge-force regularizer to safely steer early denoising, while dynamically activating exact PDE residual losses only when predictions approach the valid solution manifold. This hybrid approach prevents the extreme nonlinearity of drift-diffusion equations from destabilizing the early, high-noise stages of generation.
\end{itemize}
To validate the geometric adaptability and physical generalization of the framework, we evaluated PCGD on a highly challenging mixed benchmark comprising both PN diodes and planar MOSFETs. Across 6,763 validation snapshots spanning distinct topologies and operation modes, PCGD successfully captures disparate device physics simultaneously without suffering from catastrophic forgetting or geometric overfitting. The iterative refinement and global conditioning mechanisms reduce the mean coupled-field relative $L_2$ error from $1.585\%$ (local diffusion without cross-attention) and $1.207\%$ (one-step direct regression) down to $0.720\%$ for pure condition-aware diffusion. Furthermore, the full PCGD objective retains a sub-percent mean error of $0.835\%$ while reducing the aggregate residual error by $17.5\%$ relative to pure diffusion. On an unseen SOI-MOSFET family, rank-8 LoRA adaptation reaches a mean error of $0.815\%$ while using 5.30$\times$ fewer target trajectories and 14.34$\times$ fewer trainable parameters than a full-fine-tuning reference that reaches $0.723\%$.

\section{Related Work}

\subsection{Machine Learning Surrogates for TCAD Simulation}

Previous research in machine-learning-based semiconductor modeling has primarily focused on compact or black-box surrogates. These models directly map terminal biases and device parameters to current--voltage ($I\text{--}V$), charge--voltage, capacitance--voltage ($C\text{--}V$), variability, or lumped performance metrics \cite{wangArtificialNeuralNetworkBased2021,tungNeuralNetworkBasedBSIM2023,zhangMachineLearningBasedDevice2023,mamunComprehensiveReviewMachine2025}. While valuable for circuit simulation and rapid design-space exploration, these terminal-level outputs do not resolve the detailed spatial fields required to inspect localized transport mechanisms or evaluate microscopic conservation laws.

To resolve these field-level device states, neural operators and physics-informed neural networks (PINNs) are employed to learn continuous representations of the internal physical fields \cite{luDeepONetLearningNonlinear2021,luDeepXDEDeepLearning2021}. By embedding the coupled drift-diffusion residuals directly into the training loss function, these models enforce physical consistency, allowing both forward and inverse device modeling \cite{caoPhysicsInformedNeuralNetworks2023,rigantiDDNetUnifiedPhysicsInformed2025}. However, these mesh-free or coordinate-based approaches often struggle with the geometric complexity of real-world devices. They face significant challenges in adapting to irregular topologies, abrupt material interfaces, and steep spatial gradients, highlighting the necessity of preserving the native unstructured TCAD mesh.

To adapt field generation to complex device topologies, graph-based surrogates have been proposed to preserve the native TCAD mesh. Fan et al. predict electrostatic potential and terminal $I\text{--}V$ characteristics on unstructured device graphs, while Jang et al. jointly estimate potential, carrier densities, current, and $I\text{--}V$ curves directly on three-dimensional meshes \cite{fanGraphAttentionNetworkBased2025,jangTCADDeviceSimulation2023}. Physics-informed graph models have also been applied to coupled electro-thermal field prediction \cite{zhangDeepLearningAcceleratedUnified2025}. These approaches extend field-level TCAD prediction to irregular meshes and multiple physical outputs. However, they are still primarily deterministic final-state regressors: converged device states are directly predicted by the network based on device structures and bias conditions, with limited mechanisms for iterative conditioning or progressive physical correction.

\subsection{Generative Modeling for PDE Solving}

Generative PDE models move beyond deterministic point prediction by learning distributions over solution fields and progressively refining samples through iterative denoising or continuous probability flows \cite{hoDenoisingDiffusionProbabilistic2020,songSCOREBASEDGENERATIVEMODELING2021}. Frameworks like PDE-Refiner demonstrate the value of iterative refinement for neural PDE prediction, while recent diffusion models extend generative simulation from structured grids to unstructured mesh graphs \cite{lippePDERefinerAchievingAccurate2023,linoLearningDistributionsComplex2025,wurthDiffusionBasedHierarchicalGraph2025,giralGraphbasedDiffusionSurrogates2026}. Physics-informed diffusion and flow-matching methods further introduce governing-equation constraints during generation. However, applying these physical constraints often exposes a severe, inherent trade-off between distributional fidelity and residual satisfaction, leading to gradient conflicts when enforcing highly stiff local physics during the noisy stages of generation \cite{bastekPhysicsInformedDiffusionModels2025,baldanPHYSICSVSDISTRIBUTIONS2026}.

Despite this progress, generative modeling of coupled multiphysics fields—particularly those exhibiting extreme stiffness and exponential nonlinearity like semiconductor transport—remains largely unaddressed. Existing studies on coupled diffusion, such as ACM-FD, M2PDE, and GenCP, primarily explore flexible probabilistic emulation or coupled inference from decoupled training data \cite{longArbitrarilyConditionedMultiFunctionalDiffusion2025,zhangM2PDECompositionalGenerative2025,gaoGenCPGENERATIVEMODELING2026}. When applied to drift-diffusion systems, the exponential coupling of variables causes traditional physics-guided training to diverge rapidly. To bridge this gap, PCGD jointly generates the device state on native TCAD meshes and stabilizes training with noise-aware PDE residuals, which effectively balances data-driven generation and stiff physical constraints.

\section{Methodology}

\begin{figure*}[t]
    \centering
    \includegraphics[width=0.96\textwidth]{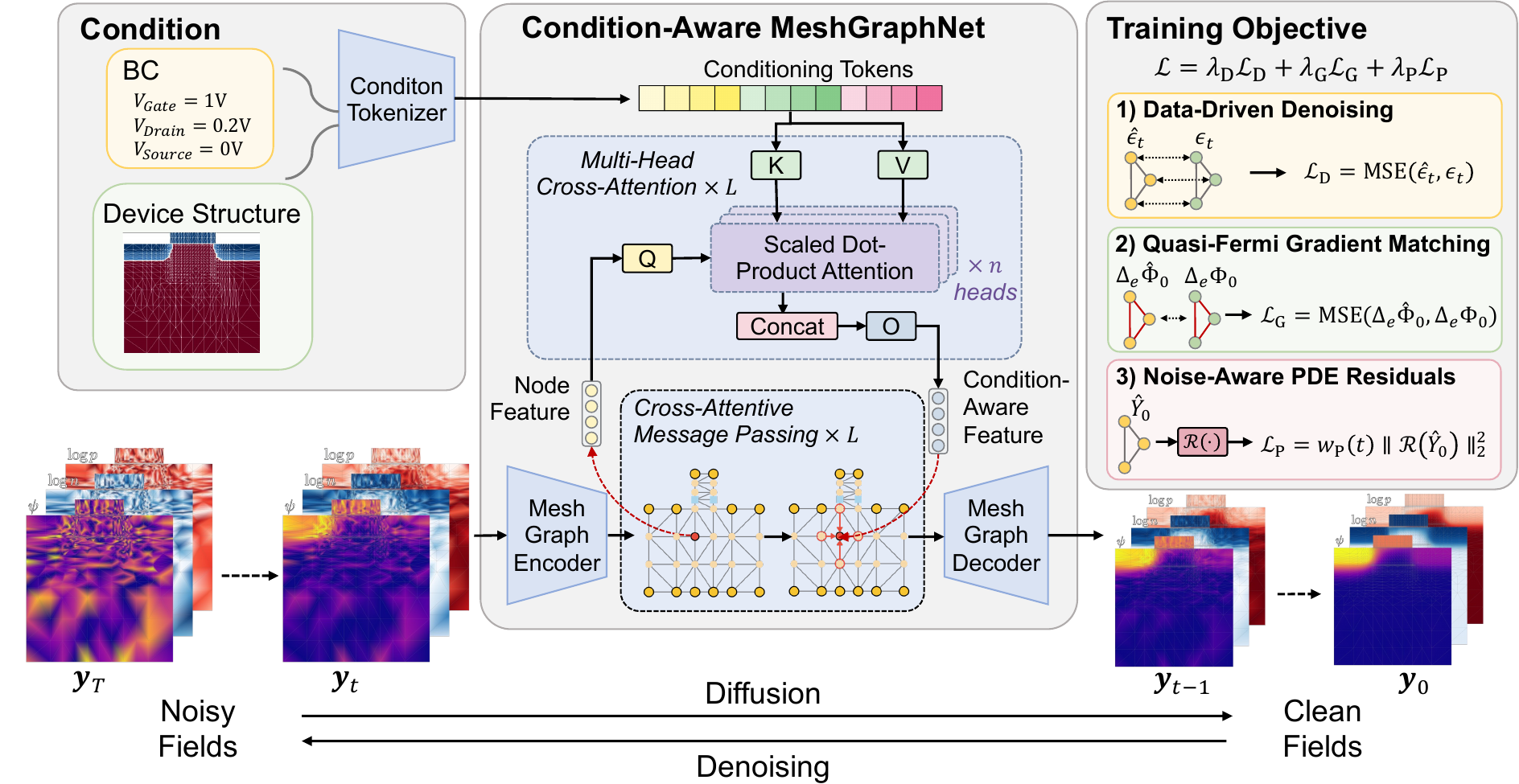}
    \caption{Overview of the PCGD framework. The condition-aware MeshGraphNet denoiser maps noisy physical fields, native TCAD mesh features, and contact-bias/device conditioning tokens to clean electrostatic and carrier-density fields, with training supervised by denoising, quasi-fermi gradient matching, and noise-aware PDE-residual objectives.}
    \label{fig:overview}
\end{figure*}

PCGD targets a field-level TCAD surrogate problem rather than scalar device-metric regression. The intended workflow replaces repeated, computationally expensive  Newton-Raphson TCAD solver for nonlinear drift-diffusion equations, while retaining the ability to route high-residual cases back to the traditional TCAD solver. Fig.~\ref{fig:overview} summarizes this mesh-native framework.

\subsection{TCAD-to-Graph Problem Formulation}

A defining challenge in applying generative AI to physical simulations is the data representation. While most existing generative models for PDEs rely on uniform Cartesian grids, industrial TCAD tools (e.g., Sentaurus) solve the drift-diffusion equations using the Finite Volume Method (FVM) on highly irregular, adaptive meshes. Rasterizing these meshes into grids destroys critical spatial resolution near interfaces and introduces interpolation artifacts.

To achieve seamless toolchain compatibility, PCGD operates directly on the native TCAD mesh. We formalize the device mesh as a spatial graph $G=(V,E)$, where $V$ represents the set of $N$ mesh nodes and $E$ represents the set of $M$ edges defining the FVM connectivity.

For each node $i \in V$, we define a static input feature vector $x_i \in \mathbb{R}^{d_v}$ encompassing local physical properties: spatial coordinates, net doping concentration, material permittivity, control volume, and a boolean contact indicator. For each edge $(i,j) \in E$, we define edge features $e_{ij} \in \mathbb{R}^{d_e}$ containing geometric properties critical for flux calculation, such as the internode direction, distance, and cross-sectional coupling area.

The physical state at any node $i$ is defined by the coupled vector of electrostatic potential and logarithmic carrier densities:
\begin{equation}
    y_i = \begin{bmatrix} \psi_i, \log n_i, \log p_i \end{bmatrix}^\top \in \mathbb{R}^3.
\end{equation}
These node-level states assemble into the global field matrix $Y \in \mathbb{R}^{N \times 3}$. Given the graph structure $G$, the global static feature matrices $X_V$ and $X_E$, and a set of global conditioning variables $c$ (which encode the terminal bias voltage and structural layout), the objective of our surrogate model is to accurately predict the target converged state, denoted as $Y_0$. This mesh-native approach avoids interpolation overhead and strictly preserves the discretization parameters exactly as the TCAD solver sees them. For multi-region devices, specialized interface edges are introduced to maintain topological connectivity; the complete schema for these features and edges is detailed in Appendix \ref{app:schema}.

\subsection{Mesh-Native Diffusion for Progressive Field Refinement}

Rather than employing a one-step deterministic regression mapping from inputs to target fields $Y_0$, we formulate this prediction as a generative task. Specifically, we model the conditional probability distribution $p_\theta(Y_0 \mid G, X_V, X_E, c)$ via an iterative denoising process \cite{hoDenoisingDiffusionProbabilistic2020}, which reconstructs the target fields through a controlled sequence of state corruption and progressive restoration.

The forward trajectory systematically obscures the converged electrostatic and carrier profiles by injecting Gaussian noise until the fields approach an uncorrelated distribution. Conversely, the reverse process acts as a sequence of conditioned refinement steps, where a neural denoiser iteratively extracts the underlying field structure from the noise. This step-by-step resolution mechanism inherently allows the model to reconstruct the solution across scales: it first recovers the macroscopic potential landscape at high noise levels, and subsequently resolves the highly localized, sharp field boundaries as the noise approaches zero.

Mathematically, the forward diffusion process progressively corrupts the clean global field state $Y_0$ over $T$ timesteps via a fixed Markov chain, adding Gaussian noise scaled by a variance schedule $\beta_1, \dots, \beta_T$:
\begin{equation}
    q(Y_t \mid Y_{t-1}) = \mathcal{N}(Y_t; \sqrt{1-\beta_t}Y_{t-1}, \beta_t \mathbf{I}).
\end{equation}

Leveraging the properties of Gaussian distributions, we can sample the noisy global state at any arbitrary timestep $t$ directly in closed form:
\begin{equation}
    Y_t = \sqrt{\bar{\alpha}_t}Y_0 + \sqrt{1-\bar{\alpha}_t}\epsilon_t, \qquad \epsilon \sim \mathcal{N}(0,\mathbf{I}),
\end{equation}
where $\alpha_t = 1 - \beta_t$ and $\bar{\alpha}_t = \prod_{s=1}^t \alpha_s$.

To invert this corruption trajectory and recover the clean fields from pure noise $Y_T \sim \mathcal{N}(0,\mathbf{I})$, we utilize a parameterized neural network to estimate the injected noise. Conditioned on the global static mesh features and the explicit device variables, this mapping is defined as: 
\begin{equation}
    \hat{\epsilon}_t = \epsilon_\theta(Y_t, t, c, X_V, X_E).
\end{equation}
The model is optimized by minimizing the expected mean squared error between the exact injected noise $\epsilon_t$ and the network's prediction across all diffusion timesteps:
\begin{equation}
    \mathcal{L}_{\mathrm{D}} = \mathbb{E}_{t, Y_0, \epsilon}\left[\|\epsilon_t - \epsilon_\theta(Y_t, t, c, X_V, X_E)\|_2^2\right].
\end{equation}
At any timestep $t$, the predicted noise $\epsilon_\theta$ enables an analytical projection of the noisy global state $Y_t$ back to an estimate of the clean target fields, denoted as $\hat{Y}_0$:
\begin{equation}
    \hat{Y}_0 = \frac{Y_t - \sqrt{1-\bar{\alpha}_t}\epsilon_\theta}{\sqrt{\bar{\alpha}_t}}.
\end{equation}
This intermediate global prediction, $\hat{Y}_0$, is the essential for integrating physics into the generative process. A structural advantage of the diffusion trajectory is that the physical validity of $\hat{Y}_0$ progressively evolves. At high noise levels (large $t$), $\hat{Y}_0$ resides far from the solution manifold, dominated by high-frequency artifacts and extreme local gradients. As denoising enters the low-noise regime (small $t$), the macroscopic field profiles stabilize into a physically meaningful state. As discussed later, this noise-to-clean evolution provides a natural "validity coordinate," dictating exactly when residual gradients become directionally informative and numerically safe to enforce.

\subsection{Condition-Aware MeshGraphNet Architecture}

The Conventional MeshGraphNet relies solely on local message passing to propagate boundary bias conditions and macroscopic device topology across a complex mesh \cite{pfaffLearningMeshBasedSimulation2021}, which suffers from severe computational inefficiency and structural information attenuation. To inject this context globally, PCGD encapsulates boundary conditions and device structure into a unified set of condition tokens $Z$.

These tokens are constructed by concatenating boundary tokens ($Z_{\mathrm{bc}}$) and structure tokens ($Z_{\mathrm{struct}}$). For each contact $k$, $Z_{\mathrm{bc},k}$ encodes its spatial coordinates, material context, and terminal bias voltages via an MLP mapping. Concurrently, $Z_{\mathrm{struct}}$ provides a fixed-length global summary of the static device topology via a query-based attention mechanism: a set of learnable device queries ($Q_{\mathrm{device}}$) attends over all projected static node features $X_V$. This mechanism allows the model to adaptively aggregate the heterogeneous mesh geometry into a compact semantic representation.

The complete condition matrix $Z$ acts as a global "semantic prompt," formed by concatenating these token sets:
\begin{equation}
    Z = \left[ Z_{\mathrm{bc}} \parallel Z_{\mathrm{struct}} \right].
\end{equation}

During the forward pass, $Z$ is injected via residual cross-attention immediately following every local message-passing block:
\begin{equation}
\begin{aligned}
    m_{ij}^{(\ell)} &= \mathrm{MLP}_e^{(\ell)}\left(h_i^{(\ell)}, h_j^{(\ell)}, e_{ij}\right),\\
    h_i^{(\ell+1)} &= \mathrm{MLP}_v^{(\ell)}\left(h_i^{(\ell)}, \sum_{j:(j,i)\in E}m_{ji}^{(\ell)}\right),
\end{aligned}
    \label{eq:message_passing}
\end{equation}
\begin{equation}
    \tilde{H}^{(\ell)} = H^{(\ell)} + \mathrm{CrossAttn}\left( \mathrm{LN}(H^{(\ell)}), \mathrm{LN}(Z), \mathrm{LN}(Z) \right).
    \label{eq:cross_attention}
\end{equation}
Here, $h_i^{(\ell)}$ represents the hidden embedding of node $i$ in layer $\ell$, $m_{ij}^{(\ell)}$ is the directed edge message, and $H^{(\ell)}$ is the dense matrix that stacks all current embeddings of nodes. $\mathrm{LN}$ denotes standard Layer Normalization. In Equation \ref{eq:cross_attention}, all mesh nodes act as queries, while the condition tokens $Z$ act as keys and values. This attention mechanism bypasses topological graph distances, granting every node direct access to terminal constraints and the structural layout, thereby substantially accelerating convergence to the correct physical operating condition.

\subsection{Hybrid Physics-Guided Training Objective}

While the denoising objective ($\mathcal{L}_{\mathrm{D}}$) successfully aligns the network with the target data distribution, it provides no structural guarantee of flux continuity. To enforce the governing physical laws of semiconductor devices, PCGD explicitly regularizes the predicted clean state $\hat{Y}_0$ through a multi-objective training scheme.

Direct enforcement of exact physical constraints is numerically challenging. Standard finite-volume TCAD evaluates carrier transport via the Scharfetter-Gummel (SG) discretization, where the electron flux along an edge $(i,j)$ relies on the Bernoulli function $B(x) = x / (\exp(x) - 1)$:
\begin{equation}
    J_{n,ij} \propto n_i B\left(-\frac{\Delta_{ij}\psi}{V_T}\right) - n_j B\left(\frac{\Delta_{ij}\psi}{V_T}\right).
\end{equation}
The exponential dependence on the normalized potential difference ($\Delta_{ij}\psi / V_T$) creates an extreme dynamic range. Evaluating these stiff exponentials on FP32 GPUs during early, noisy diffusion stages invariably induces numerical instability and gradient explosion.

To address this issue, we propose a hybrid physics-guided objective: a stable, exponent-free transport proxy to safely steer early denoising, complemented by dynamically stabilized noise-aware exact PDE residuals for fine-grained physical correction.

First, we introduce an exponent-free regularizer termed "quasi-fermi gradient matching" ($\mathcal{L}_{\mathrm{G}}$).
As shown in Appendix~\ref{app:sg}, learning the highly nonlinear SG current mathematically reduces to matching the normalized quasi-Fermi potential gradients.
For any global state $Y$, we define these current driving forces along the edge $e_{ij}$ as:
\begin{equation}
\begin{aligned}
    \Phi_{\psi}(Y,e_{ij}) &= \frac{\Delta_{ij}\psi}{V_T},\\
    \Phi_n(Y,e_{ij}) &= \ln(10)\Delta_{ij}\log n - \frac{\Delta_{ij}\psi}{V_T},\\
    \Phi_p(Y,e_{ij}) &= \ln(10)\Delta_{ij}\log p + \frac{\Delta_{ij}\psi}{V_T}.
\end{aligned}
    \label{eq:edge_force}
\end{equation}
The quasi-Fermi gradient matching loss $\mathcal{L}_{\mathrm{G}}$ is explicitly formulated as the cumulative squared error of these transport driving forces across all mesh edges, computed between the predicted field $\hat{Y}_0$ and the ground-truth target field $Y_0$:
\begin{equation}
    \mathcal{L}_{\mathrm{G}} = \frac{1}{|E|} \sum_{(i,j) \in E} \sum_{r \in \{\psi, n, p\}} \left( \Phi_r(\hat{Y}_0, e_{ij}) - \Phi_r(Y_0, e_{ij}) \right)^2.
    \label{eq:loss_gradient}
\end{equation}
Minimizing $\mathcal{L}_{\mathrm{G}}$ allows the model to safely capture fundamental transport dynamics across all denoising stages without computing stiff exponentials.

Second, exact finite-volume conservation requires the full discrete governing equations. PCGD utilizes GPU-accelerated graph operators (Appendix~\ref{app:pde_ops}) to evaluate the exact PDE residuals—Poisson ($R_{\psi}$), electron ($R_n$) and hole ($R_p$) continuity—directly on $\hat{Y}_0$:
\begin{equation}
\begin{aligned}
    \mathcal{L}_{\mathrm{P}}(\hat{Y}_0,c) &= \lambda_{\psi}\|R_{\psi}(\hat{Y}_0,c)\|_2^2 + \lambda_n\|R_n(\hat{Y}_0,c)\|_2^2\\
    &\quad + \lambda_p\|R_p(\hat{Y}_0,c)\|_2^2.
\end{aligned}
    \label{eq:phys_loss}
\end{equation}

To safely optimize these highly nonlinear exact residuals, we implement a dual-tier stabilization strategy:

\textbf{1) Validity-Aware SNR Gating ($w_{\mathrm{P}}$):} We regulate $\mathcal{L}_{\mathrm{P}}$ using the diffusion signal-to-noise ratio, clamped by a safety threshold $\gamma_{\mathrm{safe}}$:
\begin{equation}
    w_{\mathrm{P}}(t) = \max \left( 0, \, 1 - \frac{\gamma_{\mathrm{safe}}}{\mathrm{SNR}(t)} \right).
\end{equation}
This smoothly deactivates the physics penalty for noisy states ($\mathrm{SNR}(t) < \gamma_{\mathrm{safe}}$), ensuring the exact residuals contribute gradients only when predictions lie within a physically valid neighborhood.

\textbf{2) Adaptive Batch Balance ($\alpha_{\mathrm{bal}}$):} Even within safe SNR regimes, the extreme stiffness of discrete PDE constraints can trigger erratic gradient spikes. To prevent batch-level residual spikes from dominating the loss, we dynamically bound the relative influence of the physics penalty to a predefined ratio $r_{\mathrm{P}}$:
\begin{equation}
 \alpha_{\mathrm{bal}} = r_{\mathrm{P}} \frac{\mathrm{sg}(\mathcal{L}_{\mathrm{D}})}{\mathrm{sg}(\mathcal{L}_{\mathrm{P}})}, 
\end{equation}
where $\mathrm{sg}(\cdot)$ is the stop-gradient operator. 

Integrating these components, our complete training objective ($\mathcal{L}_{\mathrm{total}}$) achieves a robust balance between data fidelity and physical consistency while effectively mitigating numerical instability:
\begin{equation}
    \mathcal{L}_{\mathrm{total}} = \lambda_{\mathrm{D}}\mathcal{L}_{\mathrm{D}} + \lambda_{\mathrm{G}}\mathcal{L}_{\mathrm{G}} + \alpha_{\mathrm{bal}}\, w_{\mathrm{P}}(t)\mathcal{L}_{\mathrm{P}}.
    \label{eq:total_loss}
\end{equation}

\section{Experimental Results}

\subsection{Experiment Setup}

\textbf{Dataset} We evaluate PCGD on a mixed sequential TCAD benchmark comprising 166{,}379 training and 6{,}763 validation snapshots from 16{,}183 distinct devices. Each trajectory comprises a sequence of converged bias snapshots from a single device's voltage sweep on its native unstructured finite-volume mesh. All train/validation splits are performed at the device-trajectory level, ensuring that no geometry or bias trajectory from a validation device appears in training. The graph input encodes physical, geometric, and terminal attributes to predict the coupled state $(\psi,\log n,\log p)$ across all nodes. To ensure robust cross-architectural generalization, the benchmark is deliberately designed with a trade-off between sample volume and graph complexity. Specifically, PN diodes constitute approximately 67\% of the dataset, providing extensive geometric and doping profile variations across diverse polarities, whereas planar MOSFETs make up the remaining 33\% and introduce significantly higher structural complexity. Planar-MOSFET meshes average 1{,}755 nodes and 10{,}106 edges, yielding roughly $3.9\times$ the density of their PN-diode equivalents (455 nodes, 2{,}571 edges). By spanning this wide spectrum of high-volume fundamental junctions and high-complexity multi-terminal structures, the benchmark rigorously tests model generality. Table~\ref{tab:dataset_summary} summarizes key parameter distributions.

\begin{table}[htbp]
\caption{Summary of the mixed sequential TCAD benchmark dataset.}
\label{tab:dataset_summary}
\centering
\scriptsize
\begin{tabular}{@{}lcc@{}}
\toprule
\textbf{Parameter} & \textbf{PN-Diode} & \textbf{Planar-MOSFET} \\
\midrule
\textbf{Dataset (Samples)} & & \\
Train snapshots & $111{,}487$ & $54{,}892$ \\
Validation snapshots & $4{,}715$ & $2{,}048$ \\
Total devices & $10{,}648$ & $5{,}535$ \\
\midrule
\textbf{Device Variants} & & \\
Types & $n$-on-$p$, $p$-on-$n$ & NMOS, PMOS \\
Bias regimes & Forward, Reverse & Output, Transfer \\
\midrule
\textbf{Geometry} & & \\
Lateral dimension $L_x$ ($\mu$m) & $0.10 - 52.8$ & -- \\
Junction depth ($\mu$m) & $2.9\times10^{-4} - 0.89$ & -- \\
Vertical dimension $L_y$ ($\mu$m) & $0.01 - 5.1$ & -- \\
Gate length $L_g$ ($\mu$m) & -- & $0.05 - 5.0$ \\
Bulk thickness $t_{\mathrm{bulk}}$ ($\mu$m) & -- & $1.0 - 10.0$ \\
Oxide thickness $t_{\mathrm{ox}}$ (nm) & -- & $2.0 - 50.0$ \\
\midrule
\textbf{Doping} & & \\
Profiles & Erfc, Gaussian, Logistic & Gaussian \\
Concentration ($\log_{10}$ cm$^{-3}$) & $13.0 - 20.5$ & $15.0 - 21.0$ \\
Length scale $\sigma_x$ ($\mu$m) & $0.005 - 0.25$ & $0.006 - 0.69$ \\
\midrule
\textbf{Meshing} & & \\
Nodes (Range) & $39 - 1{,}709$ & $1{,}008 - 2{,}876$ \\
Nodes (Mean) & $455$ & $1{,}755$ \\
Edges (Range) & $158 - 10{,}018$ & $5{,}724 - 16{,}680$ \\
Edges (Mean) & $2{,}571$ & $10{,}106$ \\
\bottomrule
\end{tabular}
\end{table}

\begin{table}[htbp]
\caption{Summary of evaluated baselines and ablation models. All variants share a parameter-matched backbone to isolate the effects of generative refinement, global conditioning, and hybrid physical supervision.}
\label{tab:model_and_training}
\centering
\resizebox{\columnwidth}{!}{%
\begin{tabular}{@{}lccccc@{}}
\toprule
\textbf{Model} & \textbf{Params} & \textbf{Method} & \textbf{\begin{tabular}{@{}c@{}}Global \\ Cross-attn\end{tabular}} & \textbf{\begin{tabular}{@{}c@{}}Gradient \\ Matching\end{tabular}} & \textbf{\begin{tabular}{@{}c@{}}PDE \\ Residual\end{tabular}} \\
\midrule
Direct Regression & 3.673M & Regression & $\checkmark$ & $\times$ & $\times$ \\
Diffusion w/o attn & 3.674M & Diffusion & $\times$ & $\times$ & $\times$ \\
Diffusion & 3.673M & Diffusion & $\checkmark$ & $\times$ & $\times$ \\
+Gradient & 3.673M & Diffusion & $\checkmark$ & $\checkmark$ & $\times$ \\
+Gradient+Phys & 3.673M & Diffusion & $\checkmark$ & $\checkmark$ & $\checkmark$ \\
\bottomrule
\end{tabular}%
}
\end{table}

\textbf{Evaluated Models \& Architecture} We benchmark five model variants to decouple the gains from deterministic regression, iterative generative refinement, global condition injection, and physics-guided supervision. The deterministic baseline, \textit{Direct Regression}, is a mesh-native one-step regressor trained on the same mixed PN/MOS TCAD dataset and evaluated using the same condition-aware MeshGraphNet. To isolate the effect of diffusion sampling without global conditioning, \textit{Diffusion w/o cross-attn} keeps the diffusion denoising backbone but replaces each cross-attention block with a parameter-matched residual node MLP that does not read terminal or device-level condition tokens. The condition-aware \textit{Diffusion} model restores cross-attention to globally inject boundary voltages and device context while relying solely on the data-driven denoising objective, thereby testing the role of global conditioning under iterative refinement. Finally, \textit{Diffusion+Gradient} adds exponent-free quasi-Fermi gradient matching, and the full \textit{PCGD} model further incorporates adaptive PDE residual supervision, isolating the accuracy--residual tradeoff introduced by physics regularization. Table~\ref{tab:model_and_training} gives a compact ablation overview, while Appendix~\ref{app:experiment_details} summarizes the training protocol at a high level.

\textbf{Measurement of fidelity and physics consistency} Solution accuracy is primarily quantified by the relative $L_2$ error averaged across the three predicted fields:
\begin{equation}
    \mathrm{RelL2}_{\mathrm{avg}} =
    \frac{1}{3}\sum_{q\in\{\psi,\log n,\log p\}}
    \frac{\|\hat{q}-q\|_2}{\|q\|_2+\epsilon},
    \label{eq:rel_l2}
\end{equation}
where $\epsilon$ is a small numerical constant. Prior to error computation, analytical Dirichlet values are strictly imposed on contact nodes to evaluate interior and near-contact fidelity under rigid terminal constraints. Furthermore, physics consistency is assessed using the graph finite-volume residual operators employed during training. We evaluate the Poisson ($R_{\psi}$), electron-continuity ($R_n$), and hole-continuity ($R_p$) residuals over their respective physical masks, reporting a log-compressed maximum residual error:
\begin{equation}
    E_r = \log\left(1+\max_i |R_r(i)|^2\right),
    \qquad r\in\{\psi,n,p\},
    \label{eq:residual_error}
\end{equation}
with the aggregate residual error defined as $E_{\mathrm{res}} = E_{\psi} + E_n + E_p$. Lower values denote superior physics adherence. The maximum-residual formulation heavily penalizes localized physical violations—such as those near critical depletion zones or oxide interfaces—even when the global mean field error remains low.

\subsection{Accuracy and Physics Consistency}

\begin{figure*}[tb]
    \centering
    \includegraphics[width=0.98\textwidth]{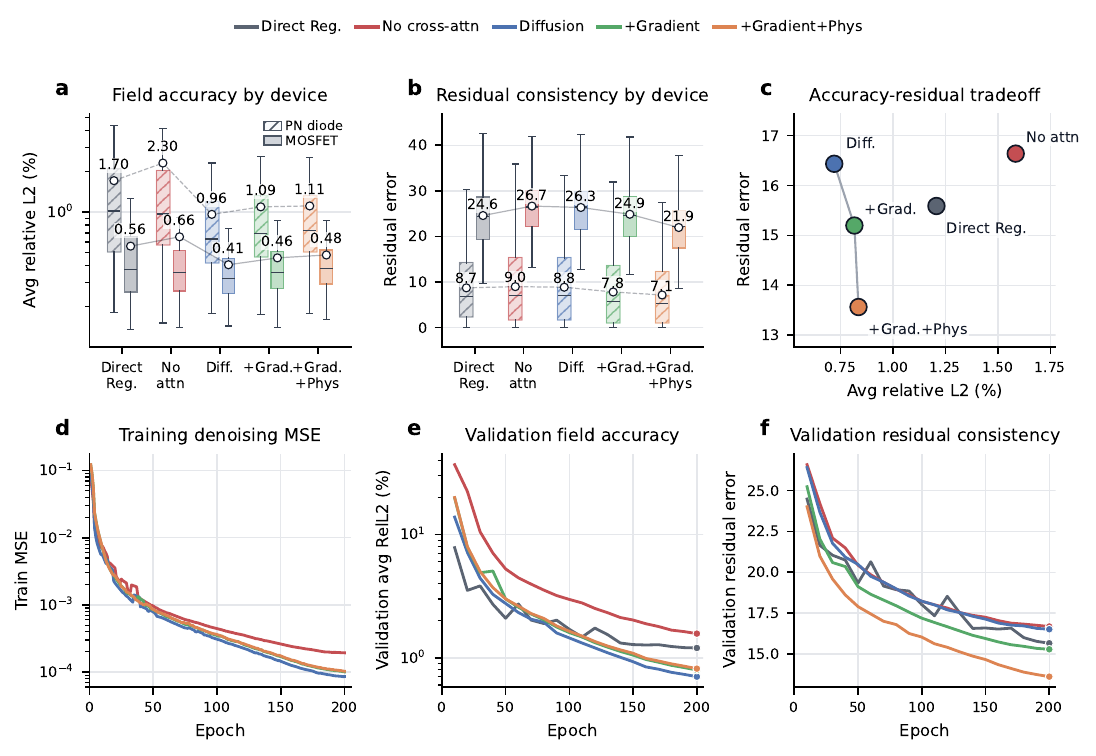}
    \caption{Main mixed-device validation and convergence comparison. (a,b) Final field accuracy and residual consistency split by PN diode and MOSFET families. (c) Accuracy--residual tradeoff among the five ablation points. (d--f) Training denoising MSE and full-sampling validation dynamics. The sequence from Direct Regression to diffusion, from local diffusion to condition-aware diffusion, and from data-only diffusion to physics-regularized PCGD isolates the contributions of iterative refinement, global conditioning, and physical regularization.}
    \label{fig:method_comparison}
\end{figure*}

\textbf{Sub-Percent Accuracy via Condition-Aware Diffusion.} 
Fig.~\ref{fig:method_comparison}a--c summarizes the validation results across 6,763 graph snapshots, firmly validating our core hypothesis that iterative generative refinement is essential for capturing highly localized, stiff physical transitions. Deterministic single-step models are prone to over-smoothing the steep gradients inherent to semiconductor interfaces and depletion zones, resulting in their relative $L_2$ error typically plateaus prematurely during optimization. In contrast, the progressive refinement of the diffusion trajectory effectively resolves these abrupt physical transitions by decoupling macroscopic field topology construction from microscopic detail refinement. Consequently, condition-aware \textit{Diffusion} significantly reduces the mean relative $L_2$ error from $1.207\%$ (\textit{Direct Regression}) to $0.720\%$, while tightening the 25th--75th percentile error interval from $0.337\%$--$1.377\%$ to $0.316\%$--$0.807\%$.

Furthermore, moving from local diffusion to condition-aware diffusion verifies the importance of global conditioning: restoring cross-attention to terminal and device tokens lowers the field error from $1.585\%$ to $0.720\%$, demonstrating that the diffusion process alone is insufficient without direct access to global operating constraints.

\textbf{Achieving Physical Consistency via Hybrid Supervision.} 
The ablation study of the three generative variants demonstrates the effectiveness of the hybrid objective in securing strict governing-equation compliance without compromising point-wise field accuracy. While pure Diffusion strictly optimizes data fidelity (achieving a $0.720\%$ field error), it lacks physical constraints, yielding the highest aggregate residual error ($16.44$). Incorporating quasi-Fermi gradient matching (\textit{Diffusion+Grad}) specifically targets the transport processes, reducing the mean electron- and hole-continuity residual errors from $5.08$ to $4.33$ and $2.70$ to $2.14$, respectively, bringing the aggregate residual down to $15.20$. Combining this quasi-Fermi gradient regularization with adaptive physics supervision (\textit{Diffusion+Grad+Phys}) provides complementary gains. This full hybrid approach further suppresses the aggregate residual error to $13.56$ while safely maintaining a sub-percent mean field error of $0.835\%$, successfully achieving both high geometric fidelity and robust physical consistency.

\begin{figure*}[t]
    \centering
    \includegraphics[width=0.85\textwidth]{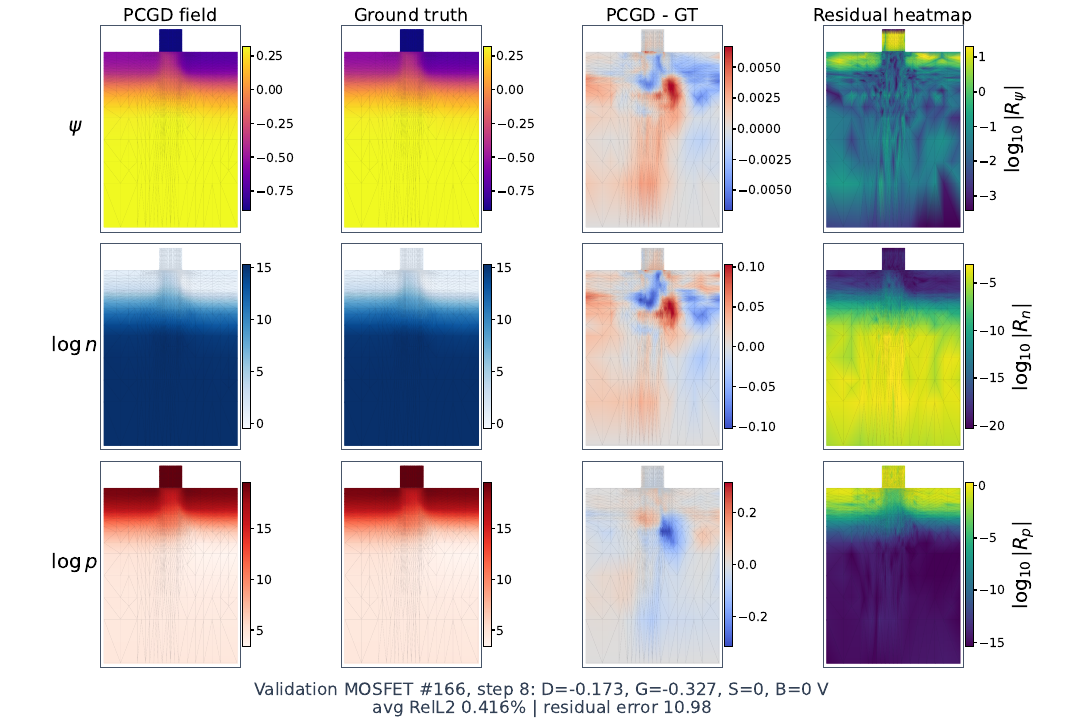}
    \caption{MOSFET coupled-field reconstruction and residual localization for the Diffusion+Grad+Phys variant. Rows correspond to $\psi$, $\log n$, and $\log p$; columns show the PCGD field, ground-truth TCAD field, signed field difference, and the corresponding $\log_{10}|R|$ residual heatmap. The residual column uses the Poisson residual $R_\psi$ for $\psi$, electron-continuity residual $R_n$ for $\log n$, and hole-continuity residual $R_p$ for $\log p$. Contact nodes are clamped to analytical Dirichlet values before residual evaluation.}
    \label{fig:mosfet_field_residuals}
\end{figure*}

Beyond aggregate metrics, the device split underscores the distinct consistency challenges inherent in multi-terminal topologies. Planar MOSFETs consistently exhibit larger residual errors than PN diodes across all diffusion variants, despite achieving comparatively lower relative $L_2$ field errors ($0.407\%$ vs. $0.960\%$ under pure \textit{Diffusion}). As visualized in Fig.~\ref{fig:mosfet_field_residuals},  PCGD successfully synthesizes the sharp boundaries of depletion regions and the extreme spatial confinement of carrier inversion channels, as evidenced by the stark transition zones in the simulated spatial profiles. This high visual fidelity, combined with deeply suppressed residual heatmaps under the hybrid objective, confirms that PCGD effectively circumvents the extreme stiffness of the drift-diffusion system to faithfully recover critical high-frequency localized physics.

\subsection{Convergence Dynamics}

As depicted in Fig.~\ref{fig:method_comparison}d--f, the optimization trajectories over 200 epochs demonstrate robust convergence under joint-distribution training. While directly enforcing highly nonlinear PDE constraints on single-step models often triggers severe gradient divergence, PCGD's iterative generative paradigm inherently circumvents these conflicts. By dynamically gating the physics penalty based on the denoising progression, the core data-driven objectives decrease consistently by approximately three orders of magnitude (from roughly $1.5\times10^{-1}$ to $1.0\times10^{-4}$) across all diffusion variants, indicating stable neural approximations and robust geometric adaptability.

Crucially, the full-sampling validation curves reveal an inherent disconnect between proxy denoising objectives and true physical consistency. While all condition-aware diffusion models converge to similar denoising MSEs, their validation residual errors diverge significantly ($16.51$, $15.29$, and $13.61$ for Diffusion, Diffusion+Gradient, and full PCGD, respectively, at epoch 200). This observation confirms that minimizing data-driven reconstruction loss is insufficient to guarantee physical conservation. Instead, quasi-Fermi gradient matching and adaptive residual supervision actively correct the sampled physical state during inference, embedding essential domain consistency that cannot be captured by the isolated forward-path training loss alone.

\subsection{Structural Adaptability to Unseen Topologies}
\label{subsec:ood}
To determine whether the pretrained PCGD representation abstracts reusable physical dynamics rather than overfitting to pretraining geometries, we evaluate its transferability to Silicon-On-Insulator (SOI) MOSFETs. This device family introduces severe out-of-distribution (OOD) topological shifts—namely, a thin silicon film, a buried-oxide (BOX) region, and a handle substrate—that are entirely absent from the pretraining dataset. Direct zero-shot transfer highlights the inherent difficulty of this geometric shift. Across the 1{,}312 held-out SOI snapshots, the model yields a high mean relative $L_2$ error of 17.90\%. This degradation is pervasive rather than isolated, as evidenced by a median error of 15.98\% and a 90th percentile (P90) error surging to 31.57\%. These metrics confirm that zero-shot extrapolation to fundamentally new material layouts remains challenging and necessitates representational realignment.

\begin{figure}[htbp]
    \centering
    \includegraphics[width=\columnwidth]{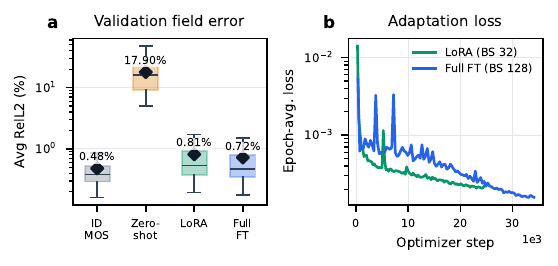}
    \caption{SOI-MOSFET adaptation. (a) Validation field-error distributions: rank-8 LoRA reaches 0.815\% mean error, approaching the 0.723\% full-fine-tuning reference while using 14.34$\times$ fewer trainable parameters. (b) Representative 100-epoch total-loss dynamics. Global batch sizes of 32 for LoRA and 128 for full fine-tuning make the optimizer-step counts appear similar, while full fine-tuning processes 5.50$\times$ more target snapshots, corresponding to approximately fivefold theoretical training time under matched utilization.}
    \label{fig:soi_transfer}
\end{figure}

While such realignment is necessary, if the network has acquired general physical priors, it should obviate the need to relearn the governing physics from scratch. To investigate this, we apply parameter-efficient fine-tuning via rank-8 LoRA adapters. Using a highly restricted adaptation set of 1{,}000 target trajectories (updating only 0.256M parameters, or 6.52\% of the model capacity), the overall mean relative $L_2$ error sharply decreases by a factor of 22$\times$ to 0.815\%. More importantly, this adaptation guarantees exceptional fidelity in typical scenarios, driving the median error down to just 0.537\%. It also demonstrates remarkable worst-case robustness: the P90 error is strictly bounded at 1.52\%, with 94.3\% of all predictions falling below a 2\% error threshold. Crucially, localized errors near device boundaries (one-hop near-contact) drop from 1.742\% (zero-shot) to 0.487\%. As depicted in Fig.~\ref{fig:soi_transfer}, this data-efficient adaptation successfully mitigates the OOD topological shift.

To benchmark this efficiency, we establish a full-parameter fine-tuning reference (updating all 3.673M base parameters using 5{,}304 trajectories). This high-cost upper bound achieves a 0.723\% mean error, with a 0.471\% median and a 1.35\% P90. The marginal performance gaps across all three distributional metrics between the parameter-efficient LoRA adaptation and the full-fine-tuning reference validate our core hypothesis: the native-mesh architecture and condition-token mechanism successfully encode fundamental, reusable physical dynamics. Consequently, when confronted with an unseen device topology, the generative surrogate requires only lightweight recalibration of structural boundaries, rather than a comprehensive retraining of the physical transport behavior.

\section{Discussion and Future Work}
\label{sec:discussion_and_future_work}
Empirical evaluations demonstrate that PCGD successfully overcomes the limitations of single-step regression models. By decoupling macroscopic field generation from microscopic physical corrections, its iterative denoising trajectory avoids the gradient conflicts typically caused by stiff PDE constraints. Furthermore, since pure data-driven generation cannot guarantee local conservation laws alone, explicit physics regularization proves essential for maintaining strict physical consistency across multi-terminal devices. Building on this validated framework, several key areas remain for future development.

\subsection{Generalization and Out-of-Distribution Extrapolation}
Efficient rank-8 LoRA adaptation on unseen Silicon-On-Insulator (SOI) MOSFETs confirms that PCGD successfully captures fundamental carrier transport and electrostatics. However, the model still exhibits severe zero-shot degradation under major topological shifts, even without introducing new transport physics. This reveals a persistent gap between structural interpolation and true physical extrapolation. To build a universal foundational model, future work must expand the pretraining corpus to encompass broader boundary conditions, material interfaces, and extreme regimes. Examples include high-voltage avalanche breakdown in PiN diodes, distinct current-control mechanisms in Bipolar Junction Transistors (BJTs), and even randomly generated, non-functional device geometries designed for the learning of universal, topology-agnostic representations. 

\subsection{Scaling to 3-D Simulation}
A primary motivation for replacing traditional TCAD solvers is computational scalability. Simulating advanced 3-D architectures (e.g., FinFETs and GAA nanosheets) requires massive, irregular meshes with drastically increased node counts ($N$) and suffers severe nonlinearities induced by quantum confinement effects and complex high-k/metal interfaces. Traditional solvers, constrained by Newton-Raphson iterations and sparse Jacobian inversions, exhibit super-linear complexity between $O(N^{1.5})$ and $O(N^2)$, imposing a severe computational bottleneck. In contrast, PCGD's message-passing formulation scales exclusively with the number of mesh edges. Given the bounded local connectivity of physical meshes, this translates directly to an asymptotic linear complexity of $O(N)$. Leveraging this intrinsic algorithmic advantage for large-scale 3-D simulation presents a compelling direction for future work.

\subsection{Hybrid AI-TCAD Simulation Workflow}
Traditional TCAD workflows rely on sequential voltage ramping, an inherently fragile process where stiff intermediate states can induce Jacobian singularities and catastrophic sweep divergence. PCGD mitigates this by generating full-field states directly from terminal conditions, decoupling from convergent steps. This independence enables a robust hybrid simulation workflow. PCGD rapidly generates AI-driven initial fields and employs GPU-accelerated operators to instantly quantify their PDE residuals. Low-residual predictions serve as high-quality initial guesses that strongly facilitate rapid TCAD convergence. Moreover, at challenging bias points where Newton-Raphson solvers typically stall, PCGD presents a promising pathway for bypassing singularities. By providing stable starting states for subsequent steps, this approach could help prevent the premature termination of voltage sweeps. This synergistic integration selectively combines AI-accelerated initialization with exact TCAD resolution, expediting large-scale design exploration while strictly preserving physical fidelity.

\section{Conclusion}

Traditional TCAD simulations are computationally expensive, and existing machine-learning surrogates often sacrifice crucial microscopic field details by relying on structured grids or scalar metrics. To overcome this, we propose PCGD, a physics-guided generative conditional graph diffusion framework that operates directly on native unstructured TCAD meshes. By integrating a condition-aware MeshGraphNet denoiser with an exponent-free edge-force regularizer and noise-aware PDE residual supervision, PCGD progressively enforces exact physical constraints while effectively circumventing the gradient divergence typical of stiff nonlinear systems. 

Empirical results on a challenging mixed PN/MOS benchmark demonstrate that PCGD overcomes the accuracy limitations of one-step regression and local diffusion models, achieving a sub-percent mean relative field error of 0.835\% while simultaneously reducing aggregate physical residuals by 17.5\% compared to pure data-driven diffusion model. Furthermore, the framework demonstrates robust physical transferability. By leveraging parameter-efficient LoRA adaptation, it successfully extrapolates to unseen SOI-MOSFET topologies, achieving a 0.815\% error while utilizing 5.30$\times$ less data and 14.34$\times$ fewer trainable parameters than full fine-tuning. Moving forward, PCGD's $O(N)$ linear asymptotic complexity and its ability to generate physically consistent initial states provide a strong foundation for scaling to massive 3-D architectures and enabling highly robust, accelerated hybrid AI-TCAD simulation workflows.

\bibliographystyle{IEEEtran}
\bibliography{references}

\appendices

\section{Experimental Protocol}
\label{app:experiment_details}

\subsection{Model Training and Fine-Tuning Protocols}
Table~\ref{tab:hyperparameters} summarizes the PCGD denoiser architecture settings and the coefficients of the physics-guided objective in Eq.~\ref{eq:total_loss}.

\begin{table}[htbp]
\caption{PCGD denoiser architecture and physics-objective parameters.}
\label{tab:hyperparameters}
\centering
\begin{tabular}{@{}lc@{}}
\toprule
\textbf{Parameter} & \textbf{Value} \\
\midrule
\multicolumn{2}{@{}l}{\textbf{Architecture (Condition-Aware MeshGraphNet)}} \\
Message-passing blocks & $16$ \\
Hidden width & $128$ \\
Normalization & LayerNorm \\
Attention heads & $4$ \\
Terminal boundary tokens & $4$ \\
Structure tokens & $16$ \\
Trainable parameters & $3.673$M \\
\midrule
\multicolumn{2}{@{}l}{\textbf{Physics Objective Parameters}} \\
Denoising weight $\lambda_{\mathrm{D}}$ & $1.0$ \\
Quasi-Fermi gradient weight $\lambda_{\mathrm{G}}$ & $2\times10^{-2}$ \\
Physics balance ratio $r_{\mathrm{P}}$ & $0.03$ \\
Physics weights $\lambda_{\psi},\lambda_n, \lambda_p$ & $1, 1, 10^4$ \\
\bottomrule
\end{tabular}
\end{table}

For the SOI-MOSFET transfer task, we apply a parameter-efficient low-rank adaptation strategy to selected conditioning and message-passing components while keeping most pretrained weights frozen. Both adaptation and full fine-tuning initialize from the same pretrained checkpoint and retain the same normalization conventions and physics-guided objectives used during pretraining so that transfer improvements reflect domain realignment rather than a change in supervision.

\subsection{Baseline Architectures}
To ensure a rigorous evaluation, the main comparison uses same-backbone controls that remove only the mechanism under test. The deterministic \textit{Direct Regression} baseline uses the same condition-aware graph backbone as PCGD but performs one-step supervised prediction from a boundary-initialized graph state to the final coupled fields, thereby isolating the benefit of iterative generative refinement from architectural capacity. The \textit{Diffusion w/o cross-attn} baseline retains the diffusion objective and boundary inpainting procedure but replaces global condition injection with parameter-matched local residual processing that does not read terminal or device-level condition tokens, specifically ablating global conditioning while preserving comparable modeling capacity.

\begin{figure*}[t]
    \centering
    \includegraphics[width=0.96\textwidth]{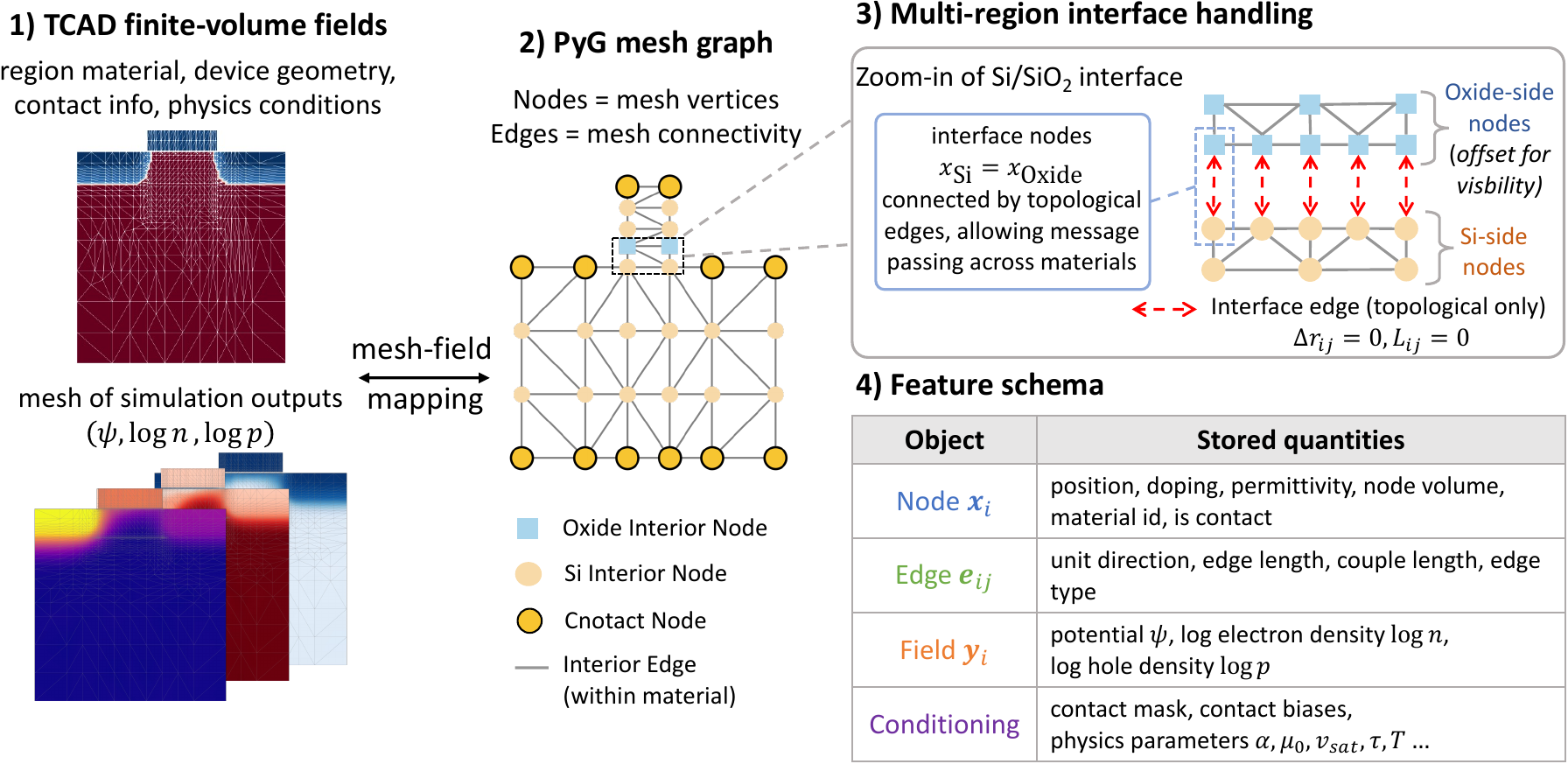}
    \caption{Mesh-native PyG representation for TCAD data. Finite-volume simulation outputs are stored directly on graph nodes and edges, with feature schemas for geometry, material, contact, field, and conditioning information. Multi-region devices preserve region-specific interface nodes and add masked topological interface edges for message passing across material regions.}
    \label{fig:pyg_graph}
\end{figure*}

\begin{figure*}[t]
    \centering
    \includegraphics[width=0.98\textwidth]{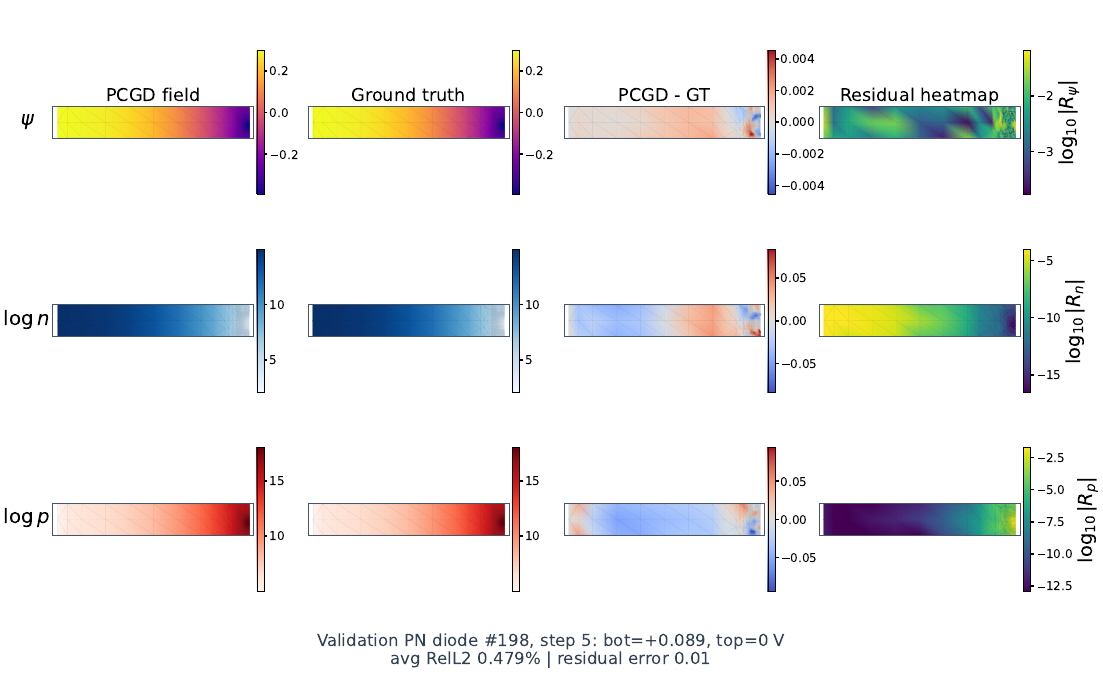}
    \caption{PN-diode coupled-field reconstruction and residual localization for Diffusion+Phys+Flux. The layout matches Fig.~\ref{fig:mosfet_field_residuals}: rows correspond to $\psi$, $\log n$, and $\log p$, while columns show PCGD, ground truth, signed difference, and $\log_{10}|R|$ residual heatmaps on the native mesh.}
    \label{fig:pn_field_residuals}
\end{figure*}

\section{Mathematical Connection Between Scharfetter-Gummel Flux and Quasi-Fermi Gradients}
\label{app:sg}

For an oriented mesh edge $e_{ij}$ (from node $j$ to $i$), let the normalized potential difference be $\eta_{ij}=\Delta_{ij}\psi/V_T$, where $\Delta_{ij}\psi=\psi_j-\psi_i$. The standard SG electron flux $J_{n,ij}$ is proportional to:
\begin{equation}
    J_{n,ij} \propto n_i B(-\eta_{ij}) - n_j B(\eta_{ij})
\end{equation}
where $B(\eta)=\eta/(\exp(\eta)-1)$ is the Bernoulli function. Utilizing the identity $B(-\eta)=\exp(\eta)B(\eta)$, we can rewrite the flux as:
\begin{equation}
    J_{n,ij} \propto B(\eta_{ij}) \left[ n_i \exp(\eta_{ij}) - n_j \right]
\end{equation}
Recognizing that the model directly predicts the base-10 logarithmic carrier density $\log n$, we substitute $n_j = n_i \exp(\ln(10)\Delta_{ij}\log n)$ into the equation:
\begin{equation}
\begin{aligned}
    J_{n,ij} &\propto B(\eta_{ij}) \left[ n_i \exp(\eta_{ij}) - n_i \exp(\ln(10)\Delta_{ij}\log n) \right] \\
    &\propto n_i B(\eta_{ij}) \exp(\eta_{ij}) \left[ 1 - \exp(\ln(10)\Delta_{ij}\log n - \eta_{ij}) \right]
\end{aligned}
\end{equation}
The term inside the inner exponent is exactly the electron quasi-Fermi gradient $\Phi_n$ defined in Eq.~\ref{eq:edge_force}. Thus, the SG electron flux can be analytically factored into:
\begin{equation}
    J_{n,ij} \propto n_i B(\eta_{ij}) \exp(\eta_{ij}) \left[ 1 - \exp(\Phi_n) \right]
\end{equation}
Following an identical derivation for the hole transport equations, the SG hole flux $J_{p,ij}$ factors as:
\begin{equation}
    J_{p,ij} \propto p_i B(\eta_{ij}) \left[ 1 - \exp(\Phi_p) \right]
\end{equation}
where $\Phi_p = \ln(10)\Delta_{ij}\log p + \eta_{ij}$.

This factorization reveals the core mathematical structure of the SG scheme: the actual driving force dictating the direction and relative magnitude of the local carrier flux is governed entirely by the $1 - \exp(\Phi)$ term. 

During the early stages of generative diffusion, evaluating the full SG current is numerically perilous. The prefactors—specifically $\exp(\eta_{ij})$ and $B(\eta_{ij})$—span extreme dynamic ranges. While traditional numerical solvers evaluate these terms safely using conservative FP64 precision on CPUs, PCGD relies on highly parallel GPU training using reduced-precision arithmetic (FP32). When the predicted potential differences ($\eta_{ij}$) are noisy, feeding them into these exponential prefactors immediately causes numerical overflow.

By analytically extracting and penalizing only the linear arguments $\Phi_n$ and $\Phi_p$ ($\mathcal{L}_{\mathrm{G}}$), PCGD directly regularizes the fundamental thermodynamic driving forces. This matches quasi-Fermi driving-force differences without exponentiation, providing a numerically stable local transport prior during high-noise generation stages.

\section{Mesh Feature and Interface-Edge Schema}
\label{app:schema}
The explicit feature representation for native finite-volume meshes is summarized in Fig.~\ref{fig:pyg_graph}. To safely handle the complexity of multi-region devices, specialized adaptations are applied to the graph topology. 

To preserve material-specific degrees of freedom (e.g., abrupt jumps in carrier concentration across a heterojunction or an oxide-semiconductor boundary), PCGD duplicates co-located mesh nodes at material interfaces, assigning each to its respective material region. To enable spatial information sharing across these hard boundaries, we introduce \textbf{topological interface edges}. These edges possess a strict geometric length vector of zero and hold no conventional finite-volume flux interpretation. A dedicated feature mask identifies them within the edge schema so that the processor blocks can utilize them exclusively for topological neural message passing, while the exact physical evaluations (quasi-Fermi gradients and PDE residual operators) safely mask them out to avoid singularity divisions.

\section{GPU-Accelerated Graph Operators for Discrete PDE Residuals}
\label{app:pde_ops}

To evaluate the physics-regularization objective $\mathcal{L}_{\mathrm{P}}$ without external solver latency, PCGD implements discrete finite-volume method (FVM) residuals as GPU-native scatter-gather operations on the mesh graph $G=(V,E)$. This formulation bypasses traditional sequential element loops by mapping FVM calculus directly onto synchronous parallel graph primitives.

\subsection{Mapping FVM Calculus to Scatter-Gather Primitives}
The structural discretization utilizes node features $X_V$ for control volumes ($V_i$) and edge features $X_E$ for cross-sectional coupling areas ($A_{ij}$) and internode distances ($L_{ij}$). Given a directed edge $(j, i) \in E$ representing flux from neighbor $j$ to target node $i$, the residual evaluation executes a two-phase tensor routing: edge-wise flux gathering followed by node-wise divergence scattering.

The discrete Poisson residual $R_{\psi, i}$ at node $i$ is formulated globally as:
\begin{equation}
    R_{\psi, i} = \sum_{j \in \mathcal{N}(i)} \varepsilon_{ij} \frac{A_{ij}}{L_{ij}} (\psi_i - \psi_j) - \frac{q V_i}{\varepsilon_0} (p_i - n_i + N_{\mathrm{net}, i})
\end{equation}
where $\mathcal{N}(i)$ excludes zero-length interface edges. Symmetrically, the electron continuity residual $R_{n, i}$ is evaluated over the graph topology as:
\begin{equation}
    R_{n, i} = q U_{\mathrm{SRH}, i} V_i - \sum_{j \in \mathcal{N}(i)} J_{n, ji}
\end{equation}
where $J_{n, ji}$ is the directed Scharfetter-Gummel electron flux along edge $j \to i$, and $U_{\mathrm{SRH}, i}$ is the Shockley-Read-Hall generation-recombination rate. For nodes residing within insulator regions, carrier concentrations and recombination rates are naturally assigned to zero.

In the GPU implementation, these domain-specific constraints are efficiently vectorized into branchless tensor operations using boolean node masks ($M_{\mathrm{semi}, i}$) and interface edge masks ($M_{\mathrm{intf}, ij}$). This eliminates conditional branching on heterogeneous hardware.

\subsection{Numerical Safety and Vectorization}
To prevent numerical instability under FP32 GPU constraints, the Scharfetter-Gummel Bernoulli function $B(x) = x / (\exp(x) - 1)$ is guarded via a localized Taylor expansion ($1 - x/2 + x^2/12$) for $|x| < 10^{-4}$ to avoid \texttt{NaN} generation. Additionally, precision truncation during mass-action law evaluations ($n_i^2$) is neutralized by isolating equilibrium carrier calculations ($n_{\mathrm{eq}}, p_{\mathrm{eq}}$) in FP64 precision before downcasting back to the core FP32 neural network backpropagation pipeline.

\end{document}